\documentclass[runningheads]{llncs}

\usepackage[T1]{fontenc}
\usepackage{multirow}
\usepackage{graphicx}

\begin{document}
\title{An analysis of data variation and bias in image-based dermatological datasets for machine learning classification}
%
%

\author{Francisco Filho\inst{1} \and Emanoel Santos\inst{1} \and Rodrigo Mota\inst{1} \and Kelvin Cunha\inst{1} \and Fábio Papais\inst{1} \and Amanda Arruda\inst{1} \and Mateus Baltazar\inst{1} \and Camila Vieira\inst{1} \and José Gabriel Tavares\inst{1} \and Rafael Barros\inst{1} \and Othon Souza\inst{1} \and Thales Bezerra\inst{1} \and Natalia Lopes\inst{1} \and Erico Medeiros\inst{1} \and Jéssica Guido\inst{2} \and Shirley Cruz\inst{2} \and Paulo Borba\inst{1} \and Tsang Ing Ren\inst{1}}
  
\institute{Centro de Informática, Universidade Federal de Pernambuco (UFPE), Pernambuco - Brazil \and Ambulatório de Dermatologia, Hospital das Clinicas - UFPE, Pernambuco - Brazil} 

\maketitle            
\begin{abstract}

AI algorithms have become valuable in aiding professionals in healthcare. The increasing confidence obtained by these models is helpful in critical decision demands. In clinical dermatology, classification models can detect malignant lesions on patients' skin using only RGB images as input. However, most learning-based methods employ data acquired from dermoscopic datasets on training, which are large and validated by a gold standard. Clinical models aim to deal with classification on users' smartphone cameras that do not contain the corresponding resolution provided by dermoscopy. It can contain captures from uncontrolled environments, skin tone variations, viewpoint changes, noises in data and labels. A possible alternative would be to use transfer learning to deal with the clinical images. However, as the number of samples is low, it can cause degradations on the model's performance. This work aims to evaluate the gap between dermoscopic and clinical samples and understand how the dataset variations impact training. It assesses the main differences between distributions that disturb the model's prediction. From experiments on different architectures, we argue how to combine the data from divergent distributions, decreasing the impact on the model's final accuracy.

\end{abstract}

\section{Introduction}

Skin cancer ranks among the neoplasias with the highest global incidence, yet few patients actively seek or have access to specialized medical attention~\cite{tsang2006even}. General practitioners often find it challenging to reach accurate conclusions, necessitating referrals to dermatology specialists for definitive diagnoses through biopsies analysis. However, biopsies are not recommended for suspected benign lesions due to their invasive nature and potential harm to the patient's skin. Additionally, the absence of a centralized structure to organize and analyze patient information, compounded by factors like physical attributes, geography, ethnicity, and age, may contribute to varying instances of skin cancer development~\cite{daneshjou2021lack,wu2021medical}. Accessibility issues, data scarcity, ethical considerations, and adherence to security and privacy rules further impede the sharing of critical patient information.

In recent years, the application of AI methods in skin cancer detection and classification has gained prominence as a valuable tool to assist medical professionals in diagnosis~\cite{freeman2020algorithm}. However, the absence of a definitive consensus on lesion severity (in the absence of biopsies) poses challenges in evaluating these approaches~\cite{wu2021medical}. The scarcity of clinical data further complicates the optimization of these methods~\cite{Daneshjou2022-cd}. Most large-scale datasets in the literature primarily consist of images from dermoscopic devices~\cite{tschandl2018ham10000,combalia2019bcn20000,mendoncca2015ph2}. Despite their quality, dermoscopic datasets suffer from distribution biases, such as variations in skin tones and class balance, rendering them inadequate for clinical evaluations~\cite{daneshjou2021lack}. On the other hand, clinical datasets~\cite{giotis2015med,Daneshjou2022-cd,Pacheco2020-mn} offer diversity in skin tone representation. However, due to their source in clinical scenarios, image quality can vary, leading to fluctuations in data distribution, including differences in illumination, blurred images, lesions out of focus, markers, and distractors. Moreover, clinical datasets are often limited in size and unsuitable for training models from scratch.

Given these considerations, it is evident that state-of-the-art models trained on dermoscopic databases struggle to generalize effectively in clinical setups, given the diverse distribution of images. Variations in device characteristics, capture protocols, and environments further compound these challenges. Even comparisons among models trained on different clinical datasets may yield undesirable results, as data variability within clinical datasets can impact performance. Consequently, this study investigates the underlying causes of these variations by analyzing features from both dermoscopic and clinical sources. Understanding the significance of dataset distribution in model training, we experiment with different dataset configurations to assess the comparability of diverse approaches in clinical scenarios, aiming to provide more reliable insights into expected model behavior.

\section{Related Works}

\subsection{Datasets in dermatology AI}

There are two main types of datasets used for skin lesion classification: clinical images (generated by common image-capturing devices)~\cite{Daneshjou2022-cd,Pacheco2020-mn,giotis2015med}, and dermoscopic images (based on an equipment consisting of a high-quality magnifying lens and an illumination) ~\cite{combalia2019bcn20000,tschandl2018ham10000,mendoncca2015ph2}. Despite the availability of various datasets in dermatology, they often come with limitations~\cite{Daneshjou2022-cd}. Many datasets are not publicly accessible, and the methods used for data curation are not always clearly outlined. Additionally, crucial demographic information such as ethnicity or skin tone may be omitted from study reports~\cite{hasan2023survey}.

The ISIC dataset\footnote{https://www.isic-archive.com/} is widely recognized as the primary resource for dermatology research. All malignant diagnoses in this dataset have been confirmed via available histopathology. Diverse Dermatology Images (DDI)~\cite{Daneshjou2022-cd} and PAD-UFES-20~\cite{Pacheco2020-mn} are collections of clinical images. Notably, DDI is the first dataset to include malignancies in dark skin tones. While PAD-UFES-20 includes images from smartphones representing six different clinical lesions. In addition, a set of metadata is associated with each image. The significance of expertly labeled data encompassing diverse skin tones and patient metadata cannot be overstated. It ensures that algorithms perform fairly across all demographic groups and ensures equitable outcomes in dermatology research and clinical applications.

\subsection{Deep Learning on dermatology classification}
Several deep learning models applied to skin lesion classification have been developed to improve early diagnosis~\cite{hasan2023survey}. Ha et al.~\cite{ha2020identifying}, the winning solution to the SIIM-ISIC Melanoma Classification Challenge, applied an ensemble of convolutional neural network (CNN) combining input images with dataset metadata, and reaches an AUC of 96\% on ISIC 2020 dataset. The ISIC 2020 dataset has only 1.76\% of positive samples (i.e., malignant) out of 33,126 images, which makes the evaluation of the model difficult. To address this problem, the ISIC 2018 and 2019 datasets were utilized.  
Sadik et al.~\cite{SADIK2023100143} proposed an analysis of different CNN architectures with transfer learning models pretrained on ImageNet, evaluating architectures such as MobileNet, Xception, InceptionV3, Inception-ResNet, and DenseNet. They used two dermoscopic dataset in the study: Dermnet images (Atopic dermatitis, Eczema, Nevus, and Herpes)~\cite{oakley2016dermnet} and HAM10000 (Melanoma images)~\cite{tschandl2018ham10000} . Augmentation techniques were applied to improve the robustness of the models. Suiçmez et al.~\cite{SUICMEZ2023104729} applied processing techniques to remove hair from dermoscopic images from the HAM10000 and ISIC 2020 datasets. Subsequently, a wavelet transform was utilized for noise removal and compression. Most works in literature employ dermoscopic dataset on evaluation~\cite{hasan2023survey}.

\subsection{Interpretability on dermatology classification}

Recently there are an effort to interpret image features that makes a model to decide whether a lesion is benign or malignant~\cite{salahuddin2022transparency}.  Attention-based approaches for melanoma recognition leverages attention maps to accentuate pertinent regions relevant to lesion classification \cite{yan2019melanoma}. Some approaches \cite{ha2020identifying} inspect outcomes from multiple networks to identify the most proficient performers. Their technique involves training various models independently and subsequently combining their outputs into an ensemble model. Despite the time-consuming nature of this method, notable achievements are achieved in terms of results.
\section{Methodology}

\subsection{Motivation}

According to Daneshjou et al.\cite{Daneshjou2022-cd}, dermatology classification approaches trained on dermoscopic datasets prove ineffective in clinical evaluations. Public datasets, primarily dermoscopic, exhibit biases and lack the necessary variation to handle real-world clinical classifications adequately. These datasets, dominated by specific characteristics, notably fall short of representing variations in skin tones and patient ethnicity.

While state-of-the-art approaches trained on datasets like HAM1000~\cite{tschandl2018ham10000} and DeepDerm perform well on dermoscopic data, they fail in classifying samples from the proposed clinical dataset (DDI). Fine-tuning the models improves their clinical classification results. However, while fine-tuning adjusts parameters for a new target within the limited dataset, other concerns related to datasets conception~\cite{Pacheco2020-mn} need consideration.

\begin{figure}[!t]
\centering
\includegraphics[width=0.4\columnwidth]{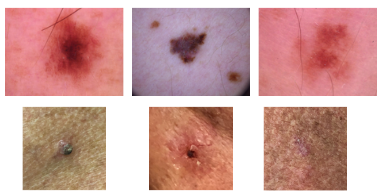}
\caption{Examples of images in the dermoscopic ISIC18 dataset (top row) and PAD-UFES-20 (bottom row). While clinical features impact model decisions, it is evident how pixel differences arise from intrinsic characteristics of each domain (e.g., capture device quality, lighting, noise, resolution).}
\label{fig:dataset_samples}
\end{figure}

The DDI dataset comprises images captured by different smartphones during patient evaluations, encompassing various skin tones, perspectives, scene illuminations, camera resolutions, and image noises. In contrast, dermoscopic datasets like HAM10000 and DeepDerm feature images captured by dermoscopic devices, providing higher resolution, controlled perspectives and illuminations, and gold-standard annotations. In Figure~\ref{fig:dataset_samples}, it is possible to observe the differences between the samples contained in the dermoscopic dataset ISIC18 and the clinical dataset PAD-UFES-20. 

\subsection{Dataset analysis}

To comprehend the features influencing model predictions, we analyzed the features extracted by deep learning models from clinical and dermoscopic datasets, considering categorical (lesions ID) and binary (malignant, benign) classification. 

Additionally, we analyzed features to visualize each lesion's characteristics, identifying differences and similarities in class boundaries that pose challenges in each classification task. Dermoscopic dataset ISIC18, including data from HAM10000~\cite{tschandl2018ham10000} and BCN20000~\cite{combalia2019bcn20000}, were used, along with the clinical dataset PAD-UFES-20~\cite{Pacheco2020-mn}. We also applied DDI~\cite{Daneshjou2022-cd} to compare results for two different clinical datasets.

\subsection{Models evaluation}
\label{sec:model_training}

Training and testing employed deep learning classification architectures, including ConvNext (Tiny, Small, and Base)~\cite{liu2022convnet}, DenseNet (121 and 161)~\cite{zhang2019multiple}, ResNet (50 and 152), and EfficientNet. Various setups were assessed to evaluate accuracy under domain variability by altering training and testing distributions:

\begin{itemize}
    \item Full Dermoscopic (\textbf{FDerm}): Models were trained on ISIC training set
    \item Full clinical (\textbf{FClinic}): Models were trained on PAD-UFES-20 training set\footnote{As PAD-UFES-20 do not contain an explicit training partition, we divided the samples equally between classes using 50\% of the data to train}
    \item Fine-tuned models (\textbf{FineClinic}): Models trained on dermoscopic data (ISIC) are fine-tuned on the clinical dataset (PAD-UFES-20), using only 30\% of the clinical samples separated for training.
\end{itemize}

Augmentations were incorporated into model training to mitigate image noise and class imbalance impacts. Transformations, such as RandomHorizontalFlip, RandomVerticalFlip, RandomRotation, ColorJitter, RandomResizedCrop, and RandomAffine, were applied to images representing a smaller percentage of the total training data. Models were implemented using the PyTorch library~\cite{pytorch2018pytorch} on a system with a 12th Gen Intel Core i7-12700H CPU, 16GB RAM, and a GPU 3060 with 12GB VRAM. Starting with pre-trained weights from IMAGENET~\cite{deng2009imagenet}, we trained each model for 100 epochs using the ADAM optimizer with a learning rate of $1 \times 10^{-1}$ and a Cosine Annealing scheduler that decreased the learning rate every 10 epochs.

\section{Results and Discussions}


\subsection{Dermoscopic and clinical models}

In the initial phase of training the model, we assessed various architectures by evaluating each model on dermoscopic and clinical data within the respective domains. Following the operations outlined in Section~\ref{sec:model_training}, augmentation procedures were applied. Table~\ref{tab:overall} provides an overview of the overall accuracy achieved by each model. While there are slight variations in the results across architectures, the classification behavior remains similar among models. The ConvNext model, yielding the best result, matches the state-of-the-art within the scope of CNN architectures.

\begin{table}[]
\caption{Results obtained from testing various model variations on the dermoscopic dataset ISIC18. Our goal was to assess the performance of different architectures in clinical dermatology evaluation. We examined the differences between training and test cases, and additionally, we compared variations in prediction scores by considering accuracy and F1-score.}
\centering
\label{tab:overall}
\begin{tabular}{cccc|cc}
\begin{tabular}[c]{@{}c@{}}Eval \\ Dataset\end{tabular} & \multicolumn{3}{c|}{Accuracy} & \multicolumn{2}{c}{F1-Score} \\ \hline
 & Train & Test & Test & Val & Test \\ \hline
ConvNext-Tiny & 99.64 & 87.56 & 83.20 & 67.44 & 72.69 \\
ConvNext-Small & 99.50 & 88.60 & 84.34 & 76.67 & 72.76 \\
ConvNext-Base & 99.8 & 91.71 & 84.32 & 88.47 & 73.91 \\
DenseNet121 & 92.81 & 78.75 & 75.59 & 63.30 & 58.48 \\
DenseNet161 & 98.68 & 84.45 & 75.00 & 80.87 & 58.84 \\
DenseNet201 & 94.04 & 86.01 & 75.31 & 61.39 & 66.54 \\
ResNet101 & 93.67 & 79.79 & 75.95 & 58.69 & 58.38 \\
ResNet152 & 84.68 & 82.38 & 81.14 & 83.63 & 59.06 \\
InceptionV3 & 98.39 & 88.60 & 80.45 & 83.63 & 67.88 \\
EfficientNet & 97.43 & 84.97 & 79.95 & 79.77 & 67.26
\end{tabular}
\end{table}

Initially, it can be stated that the models are adequate, demonstrating reasonable accuracy values. However, a closer examination of each class's performance reveals the impact of biases stemming from sample imbalances (see Table~\ref{tab:class_perf}). Most predictions are directed towards nevi, constituting over 90\% of the available samples in ISIC. The number of available samples for malignant lesions is not equivalent to benign cases, posing a challenge in obtaining a balanced dataset for malignancy. Class accuracy for malignant lesions is 59\%, 38\%, and 84\% for melanoma, SCC, and BCC, respectively. Even after training the model on the clinical dataset, it is evident that they are not suitable for real clinical applications. The model tends to prioritize predictions based on the majority class distribution, leading to many false negatives. In the ISIC dataset, we have 6705 nevi samples compared to 1113 melanoma examples. Conversely, in the PAD dataset, only 66 nevi and 11 melanoma examples exist. Thus, reports on overall dataset scores must be carefully checked, using appropriate metrics to assess class differences.

\begin{table*}[]
\caption{Accuracy results for each class were obtained by training the model on their corresponding dataset. While the average model accuracy is acceptable, individual class scores for malignant samples are inadequate for clinical evaluation. Missing values indicate that the dataset does not contain the respective class. }
\centering
\label{tab:class_perf}
\begin{tabular}{cccccccccc}
Dataset & Nevus & Melanome & \begin{tabular}[c]{@{}c@{}}Seborrheic\\ Keratosis\end{tabular} & \begin{tabular}[c]{@{}c@{}}Benign\\ Keratosis\end{tabular} & \begin{tabular}[c]{@{}c@{}}Dermato-\\ fibroma\end{tabular} & SCC & BCC & \begin{tabular}[c]{@{}c@{}}Vascular \\ Lesion\end{tabular} & \begin{tabular}[c]{@{}c@{}}Akinc.\\ Keratosis\end{tabular} \\ \hline
ISIC18 & 0.96 & 0.59 & - & 0.69 & 0.57 & - & 0.83 & 0.66 & 0.6 \\
PAD & 0.87 & 0.37 & 0.77 & - & - & 0.38 & 0.84 & - & 0.75 
\end{tabular}
\end{table*}


\subsection{Model adaptation}

\begin{table}[]
\caption{Results were obtained by adapting models through a fine-tuning strategy. The fine-tuned model is initialized on the dermoscopic dataset and adjusted using a subset of the training samples from the clinical dataset (specifically, 30\%). The models underwent training on the ISIC18 dermoscopic dataset and fine-tuning on PAD-UFES-20 samples. For simplicity, the average values from the tested models are considered.}
\centering
\label{tab:fine-tuning}
\begin{tabular}{ccccccc}
\begin{tabular}[c]{@{}c@{}}Eval \\ Dataset\end{tabular} & \multicolumn{2}{c}{\begin{tabular}[c]{@{}c@{}}FDerm\\ ISIC18\end{tabular}} & \multicolumn{2}{c}{\begin{tabular}[c]{@{}c@{}}FClinc\\ PAD\end{tabular}} & \multicolumn{2}{c}{\begin{tabular}[c]{@{}c@{}}FineClinic\\ ISIC18+PAD\end{tabular}} \\ \hline
 & ACC & F1 & ACC & F1 & ACC & F1 \\ \hline
ISIC18 & 84.34 & 73.92 & 27.53 & 21.73 & 65.54 & 42.48 \\
PAD & 22.07 & 05.22 & 80.95 & 68.95 & 75.03 & 67.18 \\
DDI & 17.53 & 13.73 & 12.80 & 11.80 & 17.68 & 11.80
\end{tabular}
\end{table}

Table~\ref{tab:fine-tuning} displays the results obtained after fine-tuning the model architectures to adapt their parameters in clinical classification. It is evident that, before adjustment, a model trained on dermoscopic datasets struggles to perform well in a clinical scenario, obtaining a 5.22 for the F1-Score. The drop in accuracy is substantial, with classification failures even in the category with the largest number of samples. However, after fine-tuning, there is a noticeable improvement in the model's performance during clinical evaluations. The F1-Score of FineClinic has shown a notable enhancement, reaching 67.18, a performance level akin to the model trained on complete clinic images. Remarkably, FineClinic achieves this comparable score while utilizing fewer real images than FClinic, underscoring its efficiency and effectiveness. Utilizing a subset of images from the available clinical dataset enhances the expected accuracy, though it may not reach the same level of performance achieved in the dermoscopic domain. This discrepancy is anticipated, given the more dispersed distribution in the clinical scenario and the limited number of available samples. It is worth mentioning that augmentations were applied during the model re-training to adjust the influence of samples, focusing on prioritizing the minority class. While there is room for improvement in this process, it reduces model bias by avoiding a presumption of only the same class.

Additionally, as shown in Table~\ref{tab:fine-tuning}, the reverse situation is true: models trained on a clinical scenario do not perform well in dermoscopic evaluations. Consequently, the variation in visual features targeted for clinical classification cannot adapt models to dermoscopic settings. Since clinical datasets encompass variations in patients and environments, they lack essential information about lesion details necessary for accurate classification by both models and medical professionals.

\subsection{Clinical to clinical evaluation}

Clinical scenarios pose a greater challenge than their dermoscopic counterparts, primarily due to the diverse variations in data distribution. However, the limited number of images also constrains optimizing the model. To explore this limitation, we conducted a comparative analysis using two distinct clinical datasets.
In this experiment, we employed a model trained on PAD-UFES-20, which encompasses the desired variations in patient skin tones, and evaluated its performance on the DDI dataset. The average results of the models tested in these scenarios are presented in the third row of Table\ref{tab:fine-tuning}.
Despite dealing with two clinical datasets, it becomes apparent that the model struggles to generalize knowledge from PAD-UFES-20 to DDI samples. Even on the same data class, these datasets exhibit distinct characteristics in terms of image resolution, lesion distance, illumination, and reflectance, lesion severity, and capture devices. Consequently, as shown in Table~\ref{tab:fine-tuning}, fine-tuning and augmentation, as outlined in the DDI source, are imperative to ensure accurate lesion predictions. Thus, in clinical scenarios, training on a limited clinical dataset cannot be guaranteed to yield a model capable of performing as desired in diverse practical applications. In light of this, it would be premature to assert that dermoscopic models are ineffective in evaluating clinical data solely due to inherent biases linked to patient demographic variations. Although this consideration is valid, it is crucial to acknowledge that other factors are concurrently impacting the model's overall performance.

\section{Conclusions}

In this study, we evaluate the characteristics of both dermoscopic and clinical datasets, exploring their features and analyzing their impact on model predictions. Through the application of samples across various CNN architectures, we assess the combination of clinical and dermoscopic data, highlighting the crucial role of domain differences in the evaluation process. Clinical features, such as patient skin tone, ethnicity, age, and lesion format, influence learning, while intrinsic factors within the domain distribution, including image resolution, distortions, noise, and illumination, also impact the model's knowledge. Additional experiments support this observation, especially when comparing two distinct clinical datasets captured in different settings. We anticipate that these experiments offer valuable insights into the application of AI in clinical dermatology. While these models prove suitable for assisting doctors in preliminary diagnoses, there remain gaps in methodological conception that need addressing for a reliable application of this technology in real-world cases. By systematically evaluating biases within each type of data, we propose alternative evaluation approaches to enhance the reliability of models in both clinical and dermoscopic setups.

%

\section*{Acknowledgement}

This project was supported by the Ministry of Science, Technology and Innovation of Brazil, with resources from Law No. 8,248, dated October 23, 1991, under the scope of the PPI-SOFTEX, coordinated by Softex and published under Residência em TIC 13, DOU 01245.010222/2022-44.

\bibliographystyle{splncs04}
\bibliography{main}

\end{document}